\documentclass[10pt,conference]{IEEEtran}

\usepackage{booktabs}
\usepackage{multirow}
\usepackage{makecell}
\usepackage{xcolor}
\usepackage{balance}
\usepackage[most]{tcolorbox}
\usepackage{tikz}
\usetikzlibrary{arrows.meta,positioning,shapes.geometric,fit,backgrounds}
\usepackage{listings}
\usepackage{enumitem}
\usepackage{pifont}
\usepackage{amsmath}
\usepackage{amssymb}
\usepackage{amsthm}
\newtheorem{definition}{Definition}
\usepackage{graphicx}
\usepackage{cite}
\usepackage{url}

\newcommand{\answerrq}[2]{\vspace{3pt}\noindent\fbox{\parbox{0.96\columnwidth}{\textbf{Answer to RQ#1:} \emph{#2}}}\vspace{3pt}}
\newcommand{\toolname}{\textsc{Hecate}}

\definecolor{promptframe}{HTML}{434343}
\definecolor{promptback}{HTML}{F4F4F4}

\newtcolorbox{promptbox}[1][]{
  colback=promptback,
  colframe=promptframe,
  fonttitle=\small\bfseries\sffamily,
  title={#1},
  boxrule=0.5pt,
  arc=2pt,
  left=3pt, right=3pt, top=1pt, bottom=1pt,
  fontupper=\small,
  before skip=3pt, after skip=3pt,
}

\begin{document}

\title{Rethinking Complexity Metrics for LLM-Integrated Applications: Beyond Source Code}

\author{
\IEEEauthorblockN{Zihao Xu\IEEEauthorrefmark{1},
Yuekang Li\IEEEauthorrefmark{1},
Gelei Deng\IEEEauthorrefmark{2},
Yi Liu\IEEEauthorrefmark{3},
Zhenchang Xing\IEEEauthorrefmark{4}\IEEEauthorrefmark{5}}
\IEEEauthorblockA{\IEEEauthorrefmark{1}University of New South Wales, Australia\\
x123201298@gmail.com, yuekang.li@unsw.edu.au}
\IEEEauthorblockA{\IEEEauthorrefmark{2}Nanyang Technological University, Singapore\\
gelei.deng@ntu.edu.sg}
\IEEEauthorblockA{\IEEEauthorrefmark{3}Griffith University, Australia\\
yi009@e.ntu.edu.sg}
\IEEEauthorblockA{\IEEEauthorrefmark{4}CSIRO's Data61, Australia \quad
\IEEEauthorrefmark{5}Australian National University, Australia\\
zhenchang.xing@anu.edu.au}
}

\maketitle

\begin{abstract}
LLM-integrated applications interleave natural language prompts with executable code. A substantial share of their behavior is governed by the prompt layer, yet all established complexity metrics target code exclusively, leaving this logic unmeasured. We introduce \toolname{}, the first tool that quantifies the complexity of such applications across both the prompt layer and the code layer. \toolname{} rests on \emph{Prompt-as-Specification}, a formalism rooted in Hoare logic that treats each prompt as a behavioral specification. Drawing on 25 complexity dimensions from published taxonomies, it produces 52 candidate metrics. We evaluate all 52 on 118 components drawn from 18 open-source repositories, using maintenance signals extracted from commit history as ground truth, and retain only those metrics that remain significant after controlling for code size. Ten of the 52 pass this filter. Seven are newly proposed metrics, each capturing distinct structural elements—such as LLM call sites, memory attributes, and prompt templates—a quality we term \emph{structural breadth} as opposed to raw volume. Among the three surviving traditional baselines, RFC captures a comparable breadth, whereas Halstead N and V persist only as a weak size residual; our strongest metrics surpass all three. The prompt-layer metrics remain significant even after additionally controlling for the strongest code metric, confirming that prompt complexity constitutes an independent dimension. In a held-out evaluation on 20 components from six previously unseen repositories, the two strongest metrics retain their predictive power, providing evidence of generalizability.
\end{abstract}


\section{Introduction}\label{sec:intro}

Large language models (LLMs) are now capable enough to follow natural-language instructions, write code, and reason through multi-step tasks.
Because of this, developers increasingly embed them inside real software, where an LLM acts as the engine that plans steps, calls tools, and decides what to do next.
The result is a fast-growing class of programs called \emph{LLM-integrated applications}, software that interleaves natural-language (NL) prompts with executable code to perform tasks such as code generation, web browsing, and multi-step planning~\cite{wang2024llmagent,xi2023rise,guo2024agents}.
Systems such as AutoGPT, MetaGPT~\cite{hong2024metagpt}, and browser-use have together drawn hundreds of thousands of GitHub stars, and frameworks such as LangChain~\cite{langchain2023} are already used in production.

As these systems move from prototypes to production, their complexity becomes a real engineering concern.
Complexity is what makes software costly, since complex programs take more effort to change, contain more defects, and are harder to debug~\cite{basili1996validation,sculley2015hidden}.
To keep this cost visible, engineers rely on complexity metrics.
Measures such as McCabe's cyclomatic complexity~\cite{mccabe1976complexity}, Halstead's software science~\cite{halstead1977elements}, and the Chidamber--Kemerer suite~\cite{chidamber1994metrics} have long acted as early-warning signals, pointing teams to the modules that most need review or refactoring~\cite{briand2000exploring}.
A good metric turns a vague sense of risk into a number a team can track.

\begin{figure}[t]
\centering
\includegraphics[width=\columnwidth]{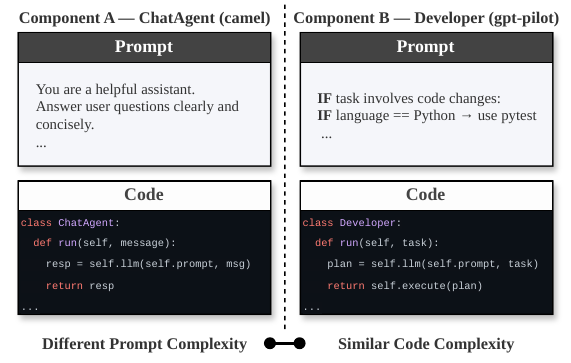}
\caption{Two components of similar code size that traditional metrics score alike but our metrics tell apart.}
\label{fig:motivating}
\end{figure}

These metrics share one assumption. They measure code, and only code.
The tools that implement them, such as Radon~\cite{radon2024} and Lizard~\cite{lizard2024}, read source and ignore everything else.
That assumption breaks for LLM-integrated applications, because much of what such an application does is not written in code at all.
It is written in the prompt.
A single prompt can carry conditional instructions such as ``if the input is ambiguous, ask for clarification'', together with role assignments, tool-routing rules, and output constraints.
This logic shapes the behavior of the application just as much as the surrounding code, yet no code metric can see it.
Figure~\ref{fig:motivating} makes the gap concrete.
Component~A and Component~B contain almost the same amount of code, but Component~B carries far more prompt logic, namely conditional rules, several tool-routing rules, and structured output contracts that Component~A lacks.
Component~B required three times as many bug-fix commits, yet a traditional metric scores the two components almost the same.
No existing complexity metric measures this hybrid of NL and code.

Building a complexity metric for these applications is hard for three reasons.
\ding{182}~\emph{Prompts have no formal structure.}
Code has well-defined representations, namely abstract syntax trees and control-flow graphs, and complexity metrics are defined on top of them.
A prompt, in contrast, is free-form text.
Before we can measure it, we need a structured model of what a prompt contains.
\ding{183}~\emph{The NL and code layers are entangled.}
The behavior of an application comes from the prompt and the code together.
Values flow from code into prompts through string interpolation, and model outputs flow back into code through parsing and tool calls.
Much of the complexity lives in this interaction, not in either layer alone.
\ding{184}~\emph{Code size dominates.}
Prior work shows that most code metrics stop predicting defects once code size is held fixed~\cite{elemam2001confounding}.
In a survey of 72 studies, ranking modules by size alone matched or beat most sophisticated prediction models~\cite{zhou2018far}.
A new metric therefore has to prove that it carries signal beyond size, rather than merely tracking it.

We present \toolname{}\footnote{Hecate is the Greek goddess of crossroads, traditionally shown with three faces looking in three directions, matching the three layers that \toolname{} analyzes, namely NL, code, and the boundary between them.}, a static-analysis tool that measures the complexity of an LLM-integrated application across both its NL and code layers.
\toolname{} meets the three challenges in turn.
For the first, it introduces \emph{Prompt-as-Specification}, a model inspired by Hoare logic~\cite{hoare1969axiomatic,meyer1992applying} that reads each prompt as a behavioral specification, namely a set of conditional rules, a set of global invariants, and the context predicates the prompt depends on.
This gives the NL layer the structure that measurement needs.
For the second, \toolname{} does not treat the two layers in isolation.
Starting from 25 complexity dimensions drawn from three published taxonomies~\cite{wang2026scaffold,complexbench2024,promptdebt2025}, it derives 52 candidate metrics that span the NL layer, the code layer, and the interface between them, so the coupling of prompt and code becomes something we can count.
For the third, \toolname{} keeps a metric only when it earns its place.
We test every candidate against real maintenance effort while holding code size fixed, and we discard any metric whose signal is explained by size alone.
\toolname{} is deterministic and dependency-free, and it runs by static analysis alone, without executing the application or calling any LLM.

We evaluate \toolname{} on 118 components from 18 open-source repositories, using maintenance signals mined from version-control history as ground truth.
The evaluation yields three findings.
First, most candidate metrics do not work.
Only 10 of the 52 keep a significant effect once code size is held fixed, namely 7 proposed metrics and 3 traditional baselines, with correlations from 0.22 to 0.40.
The seven proposed metrics measure what we call \emph{structural breadth}, since they count how many \emph{distinct} elements an application manages, such as LLM call sites, memory attributes, or prompt templates, rather than how large the application is.
The three surviving traditional baselines are weaker, and our strongest metrics outperform them.
The other 42 fail, and their failures explain why so many intuitive complexity measures are really size in disguise.
Second, the proposed metrics outperform traditional survivors.
McCabe CC loses significance once size is controlled ($\rho$ = +0.06).
RFC, a coupling metric that counts distinct reachable methods, is the strongest traditional survivor (+0.30), but our top two metrics outperform it (n\_mem\_refs +0.40, n\_llm\_calls +0.38).
The NL-layer metrics survive even after we additionally control for the strongest code metric, which shows that prompt complexity is a genuinely separate dimension rather than a byproduct of the code.
The very same idea of counting decision branches scores avg $\rho$ = +0.06 in code but +0.27 in the prompt.
Third, the findings hold beyond the studied repositories.
On 20 held-out components from 6 unseen repositories, the two strongest metrics keep their effect sizes, while the traditional baselines again fall away.

This paper makes three contributions.
\begin{itemize}[leftmargin=*]
\item \emph{An important and hard problem.}
To the best of our knowledge, we are the first to measure software complexity beyond code for LLM-integrated applications, a setting where much of the logic lives in NL prompts.
We frame the problem, spell out its three challenges, and give it a formal footing through Prompt-as-Specification.

\item \emph{A tool and a metric-design method.}
We build \toolname{}, a static analyzer that measures complexity across the NL and code layers, together with a dimension-driven method that derives 52 candidate metrics from three published taxonomies and validates each one under size control.

\item \emph{Validated and generalizable results.}
On 118 components with version-control ground truth, we identify 10 metrics that carry signal beyond code size, explain why the other 42 do not, and show that the strongest metrics still hold on unseen repositories.
\end{itemize}

\section{Background}\label{sec:background}

\subsection{LLM-Integrated Applications}

An LLM-integrated application is a software system that uses a large language model at its core, augmented with tool invocations, memory, and planning~\cite{wang2024llmagent,xi2023rise}.
Architecturally, such an application binds a \emph{prompt} (NL instructions that define role, constraints, and decision logic) to a set of \emph{tools} (executable functions the LLM may call) and an execution loop that alternates between LLM inference and tool execution~\cite{guo2024agents}.
In our study, the unit of analysis is a single programmatic construct (typically a class or a constructor call) that binds a prompt to tools. We refer to each such unit as a \emph{component} of an LLM-integrated application.
Frameworks such as LangChain~\cite{langchain2023} and MetaGPT~\cite{hong2024metagpt} have driven rapid adoption. The 18 repositories we study average over 30{,}000 GitHub stars and 2{,}700 commits.

\subsection{Software Complexity Metrics}

McCabe's cyclomatic complexity (CC)~\cite{mccabe1976complexity} counts the number of linearly independent paths through the control-flow graph of a program and remains the most widely used code complexity metric.
Halstead's software science~\cite{halstead1977elements} defines volume, difficulty, and effort from operator--operand counts.
The Chidamber--Kemerer (CK) suite~\cite{chidamber1994metrics} introduced object-oriented metrics including WMC (weighted methods per class), CBO (coupling between objects), and LCOM (lack of cohesion), validated as defect predictors by Basili et al.~\cite{basili1996validation} and Briand et al.~\cite{briand2000exploring}.
Li and Henry~\cite{lihenry1993oo} proposed NOA (number of attributes) as a maintainability predictor.
All of these metrics operate exclusively on source code and cannot capture logic expressed in NL prompts.
Moreover, even within the code layer, their predictive value has been questioned.

\subsection{The Size Confound}\label{sec:size-confound}

El Emam et al.~\cite{elemam2001confounding} showed that most CK metrics lose their association with fault-proneness after controlling for class size, concluding that size is a major confounder.
Subramanyam and Krishnan~\cite{subramanyam2003empirical} confirmed that WMC becomes statistically insignificant after size control.
Gyim\'othy et al.~\cite{gyimothy2005empirical} and Zhou and Leung~\cite{zhou2006empirical} replicated the finding on open-source Java systems.
At a larger scale, Zhou et al.~\cite{zhou2018far} surveyed 72 cross-project defect prediction studies and demonstrated that \emph{ManualDown}, a trivial model ranking modules by LOC, matches or outperforms most sophisticated prediction models.
These results establish partial correlation controlling for size as the gold standard for metric validation.
Partial correlation measures the association between two variables (a metric and a maintenance outcome) after statistically removing the effect of a third variable (the confound, here code size).
If the association of a metric vanishes after this removal, the metric is, in El Emam et al.'s framing, ``effectively a proxy for size.''
Following standard practice, we apply partial correlation to the log-transformed LOC (log(LOC)), which normalizes the right-skewed size distribution~\cite{elemam2001confounding}.
With the size confound addressed, the remaining challenge is how to formalize the NL layer so that metrics can be defined on it.

\subsection{Hoare Logic and NL Specifications}

Hoare logic~\cite{hoare1969axiomatic} reasons about program correctness through triples consisting of a precondition, a program statement, and a postcondition. Meyer's Design-by-Contract~\cite{meyer1992applying} extends this idea with preconditions, postconditions, and class invariants. More recently, Bouras et al.~\cite{bouras2025hoareprompt} use natural-language state descriptions as pre/postconditions for code verification. We instead treat prompts as behavioral specifications. This Prompt-as-Specification view provides a foundation for defining complexity metrics over prompt structure, code structure, and their interactions.

\textit{Scope of this work}. We measure \emph{software complexity}, not prompt difficulty. The distinction mirrors that between software complexity and algorithmic complexity: the former concerns the structure and maintainability of an implementation, whereas the latter concerns the intrinsic difficulty of the problem being solved. For instance, a prompt asking an LLM to prove the Riemann Hypothesis is difficult but not structurally complex. In contrast, a prompt with numerous conditional behaviors, tool interactions, and memory dependencies may be straightforward for the LLM yet difficult for developers to understand and maintain. Our metrics therefore target prompt structure rather than reasoning difficulty.

\section{The \toolname{} Approach}\label{sec:approach}

To solve this problem, \toolname{} proceeds in two stages.
First, we formalize the NL layer via Prompt-as-Specification to enable principled measurement (\S\ref{sec:formal-model}).
Second, we systematically derive 52 candidate metrics from published taxonomies (\S\ref{sec:dimensions}--\ref{sec:metric-design}).
We then implement \toolname{} as a static extraction pipeline and evaluate the metrics against version-control ground truth (\S\ref{sec:eval}).

\subsection{Prompt-as-Specification Model}\label{sec:formal-model}

To address challenge~\ding{182}, we need a way to talk about the structure of a prompt before we can measure it.
We observe that a component prompt is not free-form prose but a behavioral specification.
It tells the component what to do under which conditions, what constraints to respect, and what context to pay attention to.
We formalize this observation as follows.

\begin{definition}[Prompt-as-Specification]
\label{def:pas}
A component prompt is modeled as three parts, namely a set of \emph{behavioral rules} (individual instructions the component should follow), a set of \emph{global invariants} (constraints that apply across all rules), and a \emph{state-predicate vocabulary} (the context predicates the component conditions on).
\end{definition}

Each behavioral rule is a natural-language analog of a Hoare triple~\cite{hoare1969axiomatic}, with a \emph{condition} (when the rule fires), an \emph{action} (what the component should do), and an \emph{output constraint} (what the result must satisfy).
A rule is \emph{unconditional} when it always applies (e.g., ``always include explanations'') and \emph{conditional} when it fires only under specific circumstances (e.g., ``if the input is ambiguous, ask for clarification'').
Global invariants are constraints that hold regardless of which rule fires, for example, ``never modify the database schema without explicit approval.''

\paragraph*{Worked example}
Consider a simplified prompt from the developer component of gpt-pilot.
\begin{quote}\small
\textit{You are a senior developer. If there are syntax errors, fix them. When tests fail, analyze the failure and propose a fix. If logic is correct but style is poor, suggest improvements. Never modify the DB schema without approval. Always include explanations.}
\end{quote}
This decomposes into three conditional rules (``if syntax errors,'' ``when tests fail,'' ``if logic correct but style poor''), two invariants (``never modify DB,'' ``always include explanations''), and four state predicates that the rules condition on (syntax errors, test results, logic correctness, code style).
The formal model provides the \emph{conceptual justification} for why P\_dec\_ratio captures complexity. It counts the fraction of conditional rules among all rules, which is the NL analog of the branch-point density of cyclomatic complexity in code.
The model does not compute metrics directly. It supplies theoretical grounding that elevates keyword-based prompt analysis from ad-hoc heuristics to principled measurement, much as Hoare triples~\cite{hoare1969axiomatic} justify rather than compute program verification.

\begin{figure}[t]
\centering
\includegraphics[width=\columnwidth]{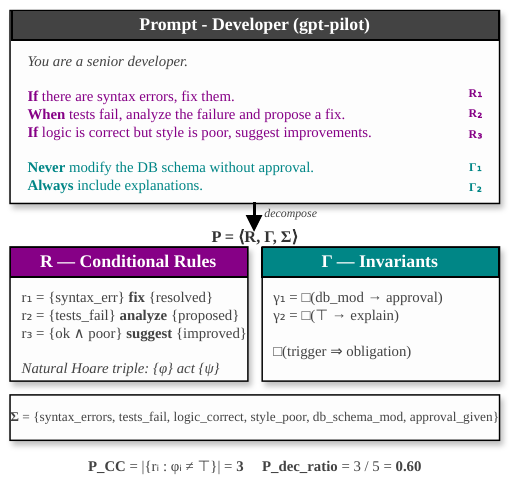}
\caption{A real component prompt decomposed into behavioral rules, invariants, and state predicates under Prompt-as-Specification.}
\label{fig:formal-model}
\end{figure}

\subsection{Dimension Identification}\label{sec:dimensions}

To address challenge~\ding{183}, we adopt a dimension-driven metric design methodology. Rather than brainstorming metrics ad hoc, we derive them systematically from published taxonomies that have empirically characterized real-world LLM-integrated applications, and we span all three layers so that the coupling between NL and code is measured rather than ignored.
We identify complexity dimensions from three sources, one per layer, then design one or more metrics per dimension.

\paragraph*{Code layer} (9 dimensions from Rombaut et al.~\cite{wang2026scaffold}).
Rombaut et al.~\cite{wang2026scaffold} conducted the first source-code analysis of 13 coding LLM-integrated applications and identified 9 architectural dimensions that distinguish them from traditional software, namely how many LLM calls the control loop composes, how many tools it binds, how it manages state across turns, and how it handles context compaction, model routing, and execution isolation.
Each of these is an architectural choice that traditional code metrics do not target, since a component with 8 LLM call sites has a fundamentally different control loop than one with 1, even if both have similar LOC.

\paragraph*{NL layer} (9 dimensions from ComplexBench~\cite{complexbench2024} and Tian et al.~\cite{promptdefects2025}).
These taxonomies characterize what makes prompt instructions complex, namely conditional branching (``if X, do Y''), constraint composition (multiple must/never rules), specification surface (how many aspects such as format, tone, and role the prompt covers), and multi-prompt scope (how many templates must stay consistent).
The key insight is that each dimension represents a \emph{specification burden}. The more conditions, constraints, and axes a prompt covers, the harder it is to maintain correctly.

\paragraph*{Code--NL interface layer} (7 dimensions from PromptDebt~\cite{promptdebt2025} and Ding and Stevens~\cite{toolintegration2025}).
These capture how tightly code and prompts are coupled, namely how many variables are injected into prompts, how LLM outputs are parsed back into code, whether tool names appear in both layers, and how changes in one layer force changes in the other.
Tighter coupling means that a change to a prompt template can break code that depends on the output format of the LLM, and vice versa.

For each dimension, we assess whether \emph{more} of it makes the application harder to maintain. Dimensions where the answer is yes become metric targets.
This yields 25 candidate dimensions across the three layers.

\subsection{Metric Design}\label{sec:metric-design}

For each of the 25 dimensions, we design candidate metrics using three strategies.

\paragraph*{Strategy A, traditional migration}
When a dimension has a conceptual analog in traditional SE metrics (McCabe~\cite{mccabe1976complexity}, Halstead~\cite{halstead1977elements}, CK~\cite{chidamber1994metrics}, Li-Henry~\cite{lihenry1993oo}), we adapt the formula to the new domain by changing the scope from code AST to NL text or from class to component.
\emph{Example.} McCabe CC counts branch points in code. P\_dec\_ratio counts the fraction of prompt instructions that contain conditional markers (``if'', ``when'', ``unless''), transposing cyclomatic density from code to NL.

\paragraph*{Strategy B, taxonomy operationalization}
When the source taxonomy describes a phenomenon qualitatively but no traditional metric exists, we turn the description into a statically computable measure.
\emph{Example.} Inside the Scaffold describes ``composable loop primitives,'' i.e., applications that compose multiple LLM reasoning steps. We turn this into n\_llm\_calls, the count of LLM invocation sites (invoke/chat/generate) in the code.

\paragraph*{Strategy C, novel design}
When neither a traditional analog nor a taxonomy formula exists, we design a new measure grounded in the core principle of the dimension.
\emph{Example.} No traditional metric measures how many distinct behavioral contracts a component manages. We define n\_prompts, the count of distinct prompt templates.

Applying these three strategies across 25 dimensions yields 52 candidate metrics. These comprise 11 traditional baselines (McCabe~\cite{mccabe1976complexity}, Halstead~\cite{halstead1977elements}, CK~\cite{chidamber1994metrics}, NOA~\cite{lihenry1993oo}) applied to code with their original definitions, plus 41 proposed metrics, namely 12 from Strategy~A (migrating traditional formulas to NL or component scope), 23 from Strategy~B (turning taxonomy descriptions into computable measures), and 6 from Strategy~C (novel designs for dimensions with no prior analog).
Not every dimension produces a metric under every strategy. A dimension receives a Strategy~A metric only if a traditional analog exists, and a Strategy~C metric only if neither A nor B applies.
The full dimension-to-metric mapping is in our supplementary material.
We describe below the 7 proposed metrics that retain significant effects after controlling for code size (\S\ref{sec:eval}). Three traditional baselines also survive and are analyzed in \S\ref{sec:rq2}.
To make the descriptions concrete, we use the \texttt{RoleZero} component of MetaGPT~\cite{hong2024metagpt} (439 LOC, 27 re-fix commits) as a running example throughout.

\paragraph*{Code layer, Strategy B (taxonomy operationalization)}

\emph{n\_mem\_refs} counts how many distinct memory-related instance attributes the component manages.
Rombaut et al.~\cite{wang2026scaffold} observe that component memory ``rang[es] from none to shared artifacts.''
We scan all \texttt{self.xxx} attributes and keep those whose name contains a memory-related word (\emph{memory}, \emph{state}, \emph{history}, \emph{context}, \emph{cache}, \emph{buffer}, \emph{log}, \emph{trajectory}, \emph{episode}).
A component with more memory channels is harder to reason about, regardless of how often each channel is accessed.
In RoleZero, four attributes match, namely \textit{self.context}, \textit{self.memory}, \textit{self.state\_machine}, and \textit{self.cmd\_prompt\_current\_state}, giving n\_mem\_refs~=~4.

\emph{n\_llm\_calls} counts how many places in the code invoke the LLM.
The same taxonomy identifies ``5 composable loop primitives'' for orchestrating multi-step reasoning.
We count call sites that follow the pattern \texttt{self.llm.chat()}, \texttt{self.model.invoke()}, or standalone calls like \texttt{aask()}.
The \texttt{self.} prefix is required to avoid false positives from unrelated methods with common names (e.g., \texttt{task.complete()}).
RoleZero invokes the LLM at 12 distinct points across its think, act, and react methods, giving n\_llm\_calls~=~12.

\paragraph*{Code layer, Strategy A (traditional migration)}

\emph{n\_attrs} adapts Li and Henry's NOA~\cite{lihenry1993oo} to component scope.
Traditional NOA~\cite{lihenry1993oo} counts attributes assigned in the constructor.
Component classes, however, configure many attributes lazily outside the constructor, so we instead count all distinct \texttt{self.xxx} references that appear anywhere in \emph{runtime methods} (excluding \texttt{\_\_init\_\_} and other dunder methods).
RoleZero references 45 distinct instance attributes across its runtime methods, giving n\_attrs~=~45.

\paragraph*{CNInterface layer, Strategy B (taxonomy operationalization)}

\emph{inject\_surf} counts how many distinct channels carry runtime values from code into prompts.
Aljohani and Do~\cite{promptdebt2025} identify ``prompt--code entanglement'' as a technical debt pattern specific to LLM applications.
We count four types of injection mechanism, namely placeholder slots in prompt strings (e.g., \texttt{\{task\}}), f-string interpolations (capped at 5 to avoid over-counting), \texttt{.format()} calls, and template renderings.
RoleZero uses 7 placeholder slots and 5 f-string injections across its 15 prompt templates, giving inject\_surf~=~12.

\paragraph*{CNInterface layer, Strategy C (novel design)}

\emph{LB} (logic balance) measures what fraction of all decision logic resides in NL prompts rather than in code.
No traditional metric captures this distribution.
The numerator counts conditional prompt instructions detected by our segmenter (\S\ref{sec:formal-model}). The denominator adds code-level branches (if, for, while, except, and boolean operators).
A higher LB means more behavior is governed by the prompt, where it is invisible to code-only tools.
RoleZero has 7 conditional prompt instructions and 57 code branches, so LB~=~7\,/\,64~=~0.11.

\paragraph*{NL layer, Strategy A (traditional migration)}

\emph{P\_dec\_ratio} transposes McCabe's cyclomatic density~\cite{mccabe1976complexity} from code to NL.
Instead of counting branch points in a control-flow graph, we measure the fraction of prompt instructions that contain conditional markers such as ``if,'' ``when,'' or ``unless.''
The density form (ratio rather than count) is naturally size-independent, which is why it retains its effect after size control while the absolute count does not.
The prompts of RoleZero contain 26 instruction units, 7 of which are conditional, giving P\_dec\_ratio~=~7\,/\,26~=~0.27.

\emph{P\_ev} adapts McCabe's essential complexity~\cite{mccabe1976complexity} to NL.
An instruction unit counts as ``essentially complex'' when it nests multiple conditions. Either two or more conditional markers appear (e.g., ``if X, unless Y''), or a conditional marker is combined with a boolean connective (e.g., ``if X \emph{and} Y'').
RoleZero has 6 such compound conditions (for example, ``If the task requires web browsing \emph{and} no browser tool is available, suggest an alternative approach''), giving P\_ev~=~6.

\paragraph*{NL layer, Strategy C (novel design)}

\emph{n\_prompts} counts the number of distinct prompt templates a component manages.
No traditional metric captures this. A component binding 15 prompt templates (system prompt, command prompt, role instruction, error prompts, etc.) manages 15 behavioral contracts whose interactions must be kept consistent.
We extract prompt strings from the source code after filtering out docstrings and short strings (less than 40 characters).
RoleZero binds 15 prompt templates for its various modes of operation, giving n\_prompts~=~15.

Table~\ref{tab:final-metrics} lists the 10 metrics that our evaluation validates, namely the 7 proposed metrics described above together with 3 surviving traditional baselines.
We define the validity test that selects them in \S\ref{sec:eval-setup} and report the complete 52-metric catalogue in our supplementary material.

\begin{table}[t]
\centering
\caption{The 10 validated metrics (7 proposed and 3 surviving traditional baselines), with Avg~$\rho$ the average partial Spearman correlation controlling for LOC over the three ground-truth dimensions.}
\label{tab:final-metrics}
\footnotesize
\setlength{\tabcolsep}{2pt}
\begin{tabular}{llllr}
\toprule
Metric & Layer & Measures & Strategy & Avg $\rho$ \\
\midrule
n\_mem\_refs & Code & \#memory-store refs & B & +0.40 \\
n\_llm\_calls & Code & \#LLM call sites & B & +0.38 \\
n\_attrs & Code & \#class attributes & A (NOA) & +0.33 \\
RFC & Traditional & Reachable method set & --- & +0.30 \\
inject\_surf & CNInterface & \#code$\to$prompt slots & B & +0.27 \\
P\_dec\_ratio & NL & Cond.\ instr.\ fraction & A (CC) & +0.27 \\
Halstead N & Traditional & Token count & --- & +0.25 \\
Halstead V & Traditional & Code volume & --- & +0.24 \\
n\_prompts & NL & \#prompt templates & C & +0.23 \\
P\_ev & NL & Essential NL conds & A (ev(g)) & +0.22 \\
\bottomrule
\end{tabular}
\end{table}

\section{Implementation and Evaluation}\label{sec:eval}

\subsection{Implementation}\label{sec:impl}

We implement \toolname{} as a fully static analysis pipeline.
Given a repository, \toolname{} identifies components through library-specific patterns (constructor names or base classes, configured for 32 libraries), resolves the \emph{footprint} of each component (the union of its code span, bound prompt constants, and tool definitions), and computes all 52 candidate metrics on the footprint.
The pipeline requires no execution, no LLM calls, and no external dependencies.
\toolname{} is publicly available as part of our replication package (see Data Availability).

\subsection{Evaluation Setup}\label{sec:eval-setup}

We evaluate \toolname{} through three research questions.
\begin{itemize}[leftmargin=*]
\item \emph{RQ1.} Which candidate metrics retain a significant effect on maintenance complexity after code size is controlled, and why do the rest fail? \emph{(Metric effectiveness and failure analysis.)}
\item \emph{RQ2.} How do the effective metrics compare with traditional complexity baselines under the same size control? \emph{(Comparison with established metrics.)}
\item \emph{RQ3.} Do the effective metrics keep their effect on unseen repositories? \emph{(Generalization beyond the study corpus.)}
\end{itemize}

\paragraph*{Corpus selection}

We searched GitHub for repositories implementing LLM-integrated applications using keywords \texttt{AI agent}, \texttt{LLM agent}, \texttt{autonomous agent}, and \texttt{agentic framework}, sorted by star count.
We screened 54 repositories and included 44 that met three criteria. First, the repository implements LLM-integrated applications with prompt-bound components. Second, each component is realized as a discrete programmatic construct (a class or constructor call binding a prompt to tools), which excludes graph-only frameworks (LangGraph), database-row components (SuperAGI), and pure-configuration paradigms (Markdown or JSON specs). Third, the repository has at least 100 commits, indicating sufficient maintenance history.
Table~\ref{tab:repos} summarizes the 18 repositories that contribute components to the final corpus.

\paragraph*{Quality filtering}

Raw extraction yielded 406 components from 29 repositories.
We applied four quality filters.
First, we removed 138 ChatDev YAML stub components (median LOC = 1), which are declarative configurations without executable code.
Second, we required LOC $\geq$ 50, removing trivial constructor calls that do not constitute meaningful component implementations.
Third, we capped each repository at 20 components to prevent single-repository dominance.
Fourth, two authors independently spot-checked 30 randomly sampled components (seed = 42) for extraction correctness, verifying that each extracted unit is indeed a component (not a utility class, test fixture, or configuration stub).
Both authors independently flagged the same 5 problematic entries (4 example-directory components and 1 false positive), which were removed. On the remaining 25 components, both authors agreed that all were correctly extracted.
The final corpus comprises 118 components from 18 repositories (median LOC = 200, range 54--1{,}902).

\paragraph*{Complexity ground truth}\label{sec:ground-truth}

Following the established tradition in metric validation~\cite{elemam2001confounding,basili1996validation}, we validate metrics against objective maintenance signals mined from version-control history rather than subjective complexity assessments.
We mine three dimensions from the footprint of each component using \texttt{git log}.

\begin{itemize}[leftmargin=*]
\item \emph{A, re-fix recurrence.}
Graves et al.~\cite{graves2000predicting} found that change history (the number of times a module has been modified) is a better predictor of future faults than any static code metric, and that modules with more past faults continue to accumulate faults, in a 1.5M-LOC telephone switching system with a formal bug-tracking database.
We adapt this to open-source repositories by classifying commit messages as bug-fixes (precision = 0.98 on manual validation) and counting elements fixed two or more times.
The underlying principle (code that required repeated correction is genuinely hard to get right) applies regardless of whether faults are tracked in a formal database or in commit messages.

\item \emph{B, temporal activity.}
Hassan~\cite{hassan2009predicting} showed that the temporal spread of code changes outperforms prior-modification count as a fault predictor across six large open-source projects.
We count the number of distinct calendar months in which the footprint of the component received an effective change.

\item \emph{C, contributor diversity.}
Bird et al.~\cite{bird2011dont} validated that the number of contributors to a component is a strong predictor of both pre-release and post-release failures.
We count distinct authors who made effective changes to the footprint of a component.
\end{itemize}

\begin{figure}[t]
\centering
\includegraphics[width=\columnwidth]{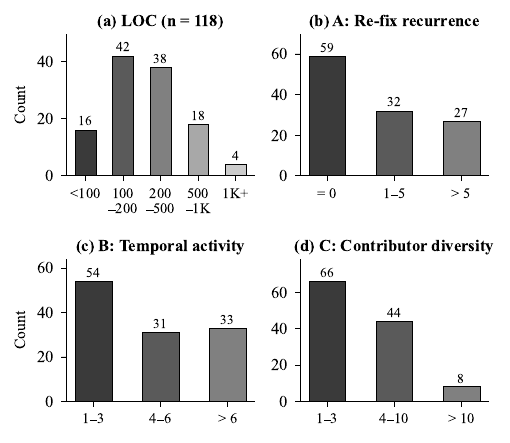}
\caption{Dataset distributions of LOC (a) and the three ground-truth complexity dimensions (b)--(d).}
\label{fig:dataset-stats}
\end{figure}

\noindent
Figure~\ref{fig:dataset-stats} shows the distributions.
Most components are small (50\% under 200 LOC), but the corpus includes four components above 1{,}000 LOC.
Dimension A is right-skewed (50\% zero, reflecting well-maintained components), while B and C are spread more evenly.

\begin{table}[t]
\centering
\caption{The 18 repositories contributing components to the corpus, typed as framework (Fwk) or end-user application (App).}
\label{tab:repos}
\small
\setlength{\tabcolsep}{3.5pt}
\begin{tabular}{llrrr}
\toprule
\textbf{Repository} & \textbf{Type} & \textbf{Stars} & \textbf{Comp.} & \textbf{Med.\ LOC} \\
\midrule
DB-GPT & Fwk & 19k & 17 & 179 \\
gpt-pilot & App & 33k & 14 & 182 \\
MetaGPT & Fwk & 68k & 10 & 276 \\
autogen & Fwk & 59k & 10 & 544 \\
letta & Fwk & 23k & 10 & 266 \\
camel & Fwk & 17k & 8 & 137 \\
aider & App & 46k & 7 & 127 \\
llama\_index & Fwk & 50k & 7 & 258 \\
DeepResearch & App & 19k & 6 & 195 \\
adk-python & Fwk & 20k & 6 & 115 \\
agent-framework & Fwk & 11k & 6 & 774 \\
agenticSeek & App & 5k & 4 & 181 \\
storm & App & 20k & 3 & 84 \\
valuecell & Fwk & 1k & 3 & 117 \\
Agent-S & App & 5k & 2 & 320 \\
gpt-engineer & App & 52k & 1 & 197 \\
smolagents & Fwk & 12k & 2 & 293 \\
SWE-agent & App & 19k & 2 & 519 \\
\midrule
Total (18 repos) & & & 118 & 200 \\
\bottomrule
\end{tabular}
\end{table}

\paragraph*{Validation protocol}
We test each candidate metric against the three ground-truth dimensions while controlling for code size.
\begin{definition}[Metric validity]
\label{def:validity}
A complexity metric is \emph{valid} for LLM-integrated applications if its association with maintenance-complexity ground truth remains statistically significant after controlling for code size (partial Spearman correlation).
\end{definition}
\noindent This criterion addresses challenge~\ding{184}, capturing El Emam et al.'s~\cite{elemam2001confounding} principle that a metric whose effect vanishes after size control is ``effectively a proxy for size.''
Following the established methodology of Basili et al.~\cite{basili1996validation}, Briand et al.~\cite{briand2000exploring}, and El Emam et al.~\cite{elemam2001confounding}, we use partial Spearman rank correlation controlling for log(LOC) as the primary evaluation criterion.
This measures the association between a metric and ground-truth complexity \emph{after removing the variance explained by code size}.
Removing the size variance this way is the accepted standard for metric validation, confirmed at scale by Zhou et al.'s 72-study survey~\cite{zhou2018far}.
Statistical significance is assessed at $\alpha = 0.05$ (two-tailed). With $n = 118$ and one control variable, the partial correlation has $n - 3 = 115$ degrees of freedom, giving a critical $|\rho| \geq 0.18$.
With 52 metrics $\times$ 3 dimensions, we report 156 tests. To guard against multiple-testing inflation, we require a metric to be significant on \emph{all three} dimensions.

\paragraph*{Sensitivity to the size threshold}
To ensure findings are not artifacts of the LOC $\geq$ 50 threshold, we re-ran the evaluation at six thresholds (0, 20, 30, 50, 100, 150), yielding sample sizes from 86 to 180 components.
All top metrics maintain direction and significance across thresholds. n\_mem\_refs avg partial $\rho$ ranges +0.35 to +0.45. P\_dec\_ratio ranges +0.06 to +0.31 (strengthening with higher thresholds, as expected when trivial components are excluded).

\subsection{Metric Effectiveness and Failure Analysis (RQ1)}\label{sec:rq1}

The majority of candidate metrics (42 out of 52) fail to provide signal beyond code size, revealing that most intuitive complexity proxies are effectively size proxies once the confound is controlled.
Ten metrics are statistically significant on all three dimensions, namely 7 proposed metrics plus the traditional RFC, Halstead~N, and Halstead~V. Eight of these measure \emph{structural breadth} rather than volume, our 7 proposed metrics together with RFC, while Halstead~N and Halstead~V survive only as a weak residual once code size is controlled.
Table~\ref{tab:full-results} reports all 52 metrics. Table~\ref{tab:final-metrics} summarizes the 10 effective ones.
Note that several metrics show \emph{negative} partial correlations. This is informative, not erroneous, since it means that engineering practices like structured output schemas and explicit constraints are associated with \emph{lower} maintenance complexity.
We organize the analysis into what works, what fails, and why.

\begin{table*}[t]
\centering
\caption{Partial $\rho \mid$ LOC for all 52 candidate metrics on the three ground-truth dimensions (n\,=\,118), with bold marking statistical significance and $\star$ marking metrics adapted from traditional literature.}
\label{tab:full-results}
\scriptsize
\setlength{\tabcolsep}{1.5pt}
\begin{tabular}{cllllllrrrr}
\toprule
\textbf{Layer} & \textbf{Metric} & \textbf{Measures} & \textbf{Example} & \textbf{Source} & \textbf{Strategy} & \textbf{Re-fix} & \textbf{Temporal} & \textbf{Authors} & \textbf{Average} \\
\midrule
Trad & RFC & Response set size & own + called methods & \cite{chidamber1994metrics} & --- & +0.32 & +0.26 & +0.31 & +0.30 \\
Trad & McCabe ev(g) & Essential branches & irreducible if-else only & \cite{mccabe1976complexity} & --- & +0.18 & $-$0.01 & $-$0.01 & +0.05 \\
Trad & Halstead E & Mental effort & volume times difficulty & \cite{halstead1977elements} & --- & +0.35 & +0.13 & +0.12 & +0.20 \\
Trad & DIT & Inheritance depth & depth of class hierarchy & \cite{chidamber1994metrics} & --- & +0.12 & +0.10 & +0.18 & +0.13 \\
Trad & Halstead N & Token count & total operators + operands & \cite{halstead1977elements} & --- & +0.35 & +0.20 & +0.19 & +0.25 \\
Trad & Halstead V & Code volume & tokens weighted by vocab & \cite{halstead1977elements} & --- & +0.34 & +0.19 & +0.18 & +0.24 \\
Trad & Halstead D & Code difficulty & operand reuse rate & \cite{halstead1977elements} & --- & +0.32 & +0.07 & +0.02 & +0.13 \\
Trad & McCabe CC & Branch-point count & if/for/while/except & \cite{mccabe1976complexity} & --- & +0.19 & $-$0.00 & +0.01 & +0.06 \\
Trad & LCOM & Cohesion lack & non-sharing method pairs & \cite{chidamber1994metrics} & --- & $-$0.16 & $-$0.02 & +0.09 & $-$0.03 \\
Trad & CBO & Coupled classes & distinct imports & \cite{chidamber1994metrics} & --- & $-$0.05 & +0.01 & +0.12 & +0.03 \\
Trad & NOA & Declared attributes & init self.x assigns & \cite{lihenry1993oo} & --- & $-$0.14 & $-$0.23 & $-$0.23 & $-$0.20 \\
\midrule
Code & n\_mem\_refs & Memory-related attributes & attrs named state/history/\ldots & \cite{wang2026scaffold} & B & +0.45 & +0.34 & +0.42 & +0.40 \\
Code & n\_llm\_calls & LLM call sites & invoke/chat/generate calls & \cite{wang2026scaffold} & B & +0.37 & +0.33 & +0.43 & +0.37 \\
Code & n\_attrs & Runtime attributes & self.xxx in non-init methods & $\star$\,\cite{lihenry1993oo} & A & +0.48 & +0.27 & +0.25 & +0.33 \\
Code & n\_sandbox & Isolation calls & subprocess/docker/sandbox & \cite{wang2026scaffold} & B & $-$0.01 & +0.09 & +0.13 & +0.07 \\
Code & n\_retry & Retry patterns & retry/backoff keywords & \cite{wang2026scaffold} & B & +0.18 & +0.00 & +0.02 & +0.07 \\
Code & n\_methods & Defined methods & methods in component & $\star$\,\cite{chidamber1994metrics} & A & $-$0.00 & +0.06 & +0.12 & +0.06 \\
Code & n\_persist & Persistence calls & save/dump/store operations & \cite{wang2026scaffold} & B & +0.00 & +0.11 & +0.03 & +0.05 \\
Code & O\_CC\_density & Branch density & branches per line of code & $\star$\,\cite{mccabe1976complexity} & A & +0.16 & $-$0.04 & +0.04 & +0.06 \\
Code & n\_callbacks & Hook registrations & callback/handler/on\_xxx & \cite{wang2026scaffold} & B & +0.09 & $-$0.05 & +0.03 & +0.02 \\
Code & n\_dispatch & Tool-dispatch branches & if-chain on tool/action name & \cite{wang2026scaffold} & B & $-$0.01 & +0.01 & +0.02 & +0.00 \\
Code & n\_tools & Tool definitions & @tool-decorated functions & \cite{wang2026scaffold} & B & $-$0.05 & $-$0.02 & +0.10 & +0.01 \\
Code & n\_parse & Output-parse calls & json.loads/parse/extract & \cite{wang2026scaffold} & B & $-$0.08 & $-$0.00 & $-$0.04 & $-$0.04 \\
Code & S\_Cx & State complexity & state vars + dict refs & \cite{wang2026scaffold} & B & $-$0.01 & $-$0.19 & $-$0.13 & $-$0.11 \\
Code & n\_mutable & Mutable state vars & attrs both written and read & \cite{wang2026scaffold} & B & $-$0.03 & $-$0.20 & $-$0.12 & $-$0.12 \\
Code & n\_models & Model references & distinct model name strings & \cite{wang2026scaffold} & B & $-$0.12 & $-$0.17 & $-$0.09 & $-$0.13 \\
Code & n\_ctx\_mgmt & Context-mgmt calls & truncate/window/compress & \cite{wang2026scaffold} & B & $-$0.01 & $-$0.23 & $-$0.18 & $-$0.14 \\
Code & n\_try\_except & Try-except blocks & error-handling blocks & \cite{wang2026scaffold} & B & $-$0.15 & $-$0.18 & $-$0.13 & $-$0.15 \\
\midrule
NL & P\_dec\_ratio & Conditional instr.\ fraction & ``if/when/unless'' share & $\star$\,\cite{mccabe1976complexity} & A & +0.25 & +0.26 & +0.29 & +0.26 \\
NL & n\_prompts & Prompt template count & distinct prompt strings & Novel & C & +0.19 & +0.25 & +0.26 & +0.23 \\
NL & P\_ev & Compound NL conditions & nested if-and/or rules & $\star$\,\cite{mccabe1976complexity} & A & +0.23 & +0.19 & +0.23 & +0.22 \\
NL & P\_V & Prompt volume & NL Halstead volume & $\star$\,\cite{halstead1977elements} & A & +0.10 & +0.01 & +0.10 & +0.07 \\
NL & P\_NOA & NL entity count & distinct capitalized terms & $\star$\,\cite{lihenry1993oo} & A & +0.12 & +0.01 & +0.11 & +0.08 \\
NL & nl\_context & Prior-context refs & ``above/previous/earlier'' & \cite{promptdefects2025} & B & +0.16 & +0.10 & +0.07 & +0.11 \\
NL & P\_V\_density & Volume per word & volume normalized by length & $\star$\,\cite{halstead1977elements} & A & +0.01 & $-$0.09 & +0.02 & $-$0.02 \\
NL & n\_constr\_broad & Constraint clauses & must/never/always count & \cite{complexbench2024} & B & $-$0.18 & $-$0.21 & $-$0.11 & $-$0.17 \\
NL & P\_LCOM & Prompt entity overlap & prompts not sharing entities & $\star$\,\cite{chidamber1994metrics} & A & +0.09 & +0.19 & +0.15 & +0.14 \\
NL & n\_specific & Quantified instructions & ``max 3''/``at most 5'' & \cite{complexbench2024} & B & +0.01 & $-$0.11 & $-$0.07 & $-$0.05 \\
NL & P\_A & Vague-term density & ``appropriate/reasonable'' rate & $\star$\,Hedging~\cite{fabbrini2001automatic} & B & +0.04 & +0.06 & +0.05 & +0.05 \\
NL & schema\_depth & Schema nesting depth & max brace nesting in prompt & \cite{complexbench2024} & B & $-$0.25 & $-$0.12 & $-$0.09 & $-$0.15 \\
NL & P\_D & Prompt difficulty & NL Halstead difficulty & $\star$\,\cite{halstead1977elements} & A & $-$0.06 & $-$0.04 & +0.03 & $-$0.03 \\
NL & spec\_surface & Specification axes & format/role/output coverage & \cite{promptdefects2025} & B & $-$0.19 & $-$0.23 & $-$0.08 & $-$0.16 \\
NL & P\_constr\_r & Constraint ratio & must/never per instruction & $\star$\,\cite{meyer1992applying} & A & $-$0.08 & $-$0.10 & $-$0.03 & $-$0.07 \\
NL & P\_RFC & Action verb count & distinct directive verbs & $\star$\,\cite{chidamber1994metrics} & A & $-$0.10 & $-$0.05 & +0.00 & $-$0.05 \\
NL & TTR & Type-token ratio & vocabulary diversity & $\star$\,Heaps' law & B & $-$0.24 & $-$0.19 & $-$0.23 & $-$0.22 \\
\midrule
CNInterface & inject\_surf & Code-to-prompt channels & slots + fstr + format + tmpl & \cite{promptdebt2025} & B & +0.36 & +0.26 & +0.18 & +0.27 \\
CNInterface & LB & Logic balance & NL decisions / all decisions & Novel & C & +0.15 & +0.16 & +0.18 & +0.16 \\
CNInterface & slot\_count & Placeholder slots & template variable count & Novel & C & +0.15 & +0.06 & +0.07 & +0.09 \\
CNInterface & shared\_funcs & Prompt-mentioned funcs & function names in both layers & \cite{toolintegration2025} & B & +0.12 & $-$0.00 & +0.12 & +0.08 \\
CNInterface & PCC & Coupling per prompt & avg variables per prompt & Novel & C & +0.12 & +0.04 & +0.03 & +0.07 \\
CNInterface & extract\_surf & Output parse points & json/regex/split extractions & \cite{wang2026scaffold} & B & +0.10 & +0.04 & $-$0.07 & +0.02 \\
CNInterface & n\_parse\_strats & Parse strategy count & distinct parse methods used & \cite{wang2026scaffold} & B & $-$0.03 & $-$0.09 & $-$0.18 & $-$0.10 \\
\bottomrule
\end{tabular}
\end{table*}

The effect sizes (0.22--0.40) are small to medium, consistent with the established finding that no single metric dominates defect prediction~\cite{basili1996validation,briand2000exploring}. What matters is significance \emph{after} size control, which most traditional metrics fail.
We now analyze why the effective metrics work and why the other 42 fail.

\paragraph*{Structural breadth}
Eight of the ten survivors count the number of \emph{distinct} entities rather than their aggregate size, namely our seven proposed metrics and the traditional RFC.
Two components of equal LOC can manage 3 or 30 memory attributes (n\_mem\_refs), invoke the LLM at 1 or 8 points (n\_llm\_calls), or embed 5\% or 50\% conditional prompt instructions (P\_dec\_ratio).
These are design choices that vary independently of code length, which is why they retain signal after size control.
The three strongest metrics (n\_mem\_refs at +0.40, n\_llm\_calls at +0.38, and n\_attrs at +0.33) all reside in the code layer but capture component-specific architectural properties (memory channels, reasoning steps, internal state) that traditional code metrics do not target.

\paragraph*{NL-layer independence}
Among the 10 survivors, three operate entirely in the NL layer, namely P\_dec\_ratio, n\_prompts, and P\_ev.
These remain significant even after additionally controlling for n\_attrs (the strongest code metric), confirming that the prompt layer carries complexity information that is invisible to code-only tools.

\paragraph*{The 42 metrics that fail}
The remaining 42 metrics yield near-zero or negative partial correlations after size control.
To understand \emph{why}, we analyzed all 42 failures and classified them into three failure modes (Table~\ref{tab:failure-modes}).

\begin{table}[t]
\centering
\caption{Failure modes of the 42 non-surviving metrics.}
\label{tab:failure-modes}
\footnotesize
\setlength{\tabcolsep}{2pt}
\begin{tabular}{@{}p{1.3cm}rp{2.1cm}p{2.1cm}p{2.0cm}@{}}
\toprule
\textbf{Mode} & \textbf{\#} & \textbf{Why it fails} & \textbf{Key insight} & \textbf{Examples} \\
\midrule
(A) Size proxy & 6 & $\rho_\text{LOC} > 0.5$, size control removes signal & Signal is code length, not design choice & P\_V, McCabe CC, S\_Cx \\[3pt]
(B) Floor effect & 4 & Near-zero variance across components & Property barely varies in real prompts & P\_A (std\,=\,0.02), P\_V\_density \\[3pt]
(C) Wrong target & 32 & Partial $\rho \leq 0$ after control & Measures style or maturity, not complexity & schema\_depth, n\_ctx\_mgmt \\
\bottomrule
\end{tabular}
\end{table}

\noindent(A) Size proxies, such as P\_V, S\_Cx, n\_methods, and O\_CC\_density, correlate with LOC at $\rho > 0.5$, so controlling for LOC removes nearly all signal.
P\_V illustrates the pattern. It correlates with prompt word count at $\rho = 0.97$, which makes it a length metric by construction.

\noindent(B) Floor effect (P\_A, P\_V\_density, P\_constr\_r, P\_D).
These metrics have extremely low variance (e.g., P\_A std = 0.02) because the measured phenomenon (vague terms, volume density) varies little across real component prompts.
With insufficient variance, no correlation can emerge.

\noindent(C) Wrong conceptual target.
Metrics such as P\_constr\_r and spec\_surface measure specification \emph{style} rather than structural complexity.
Notably, spec\_surface shows a \emph{negative} association ($\rho = -0.17$). Components that cover more specification axes (format, role, error handling) tend to be better-constrained and simpler to maintain.

\answerrq{1}{Only 10 of the 52 candidate metrics remain significant after controlling for code size. The strongest metrics measure structural breadth rather than artifact volume, while the remaining metrics fail because they are size proxies, have little variance, or target engineering style instead of complexity. These findings suggest a simple design principle: \emph{measure what is distinct, not what is large}.}

\subsection{Comparison with Traditional Baselines (RQ2)}\label{sec:rq2}

We compare the 7 proposed metrics of \toolname{} against the 11 traditional baselines, reporting both raw Spearman $\rho$ (without size control) and partial $\rho \mid \text{LOC}$ (with size control), averaged over dimensions A, B, and C.
Table~\ref{tab:comparison} shows the traditional metrics that retain the strongest partial effects alongside all 7 proposed survivors.

\begin{table}[t]
\centering
\caption{Effect of size control, comparing raw $\rho$ with partial $\rho \mid$ LOC (averaged over A, B, C) and the signal retained after control.}
\label{tab:comparison}
\small
\begin{tabular}{llrrr}
\toprule
\textbf{Metric} & \textbf{Layer} & \textbf{Raw $\rho$} & \textbf{$\rho \mid$ LOC} & \textbf{Retain} \\
\midrule
McCabe CC & Traditional & +0.39 & +0.06 & 15\% \\
Halstead D & Traditional & +0.41 & +0.13 & 32\% \\
Halstead E & Traditional & +0.45 & +0.20 & 44\% \\
Halstead V & Traditional & +0.47 & +0.24 & 51\% \\
Halstead N & Traditional & +0.47 & +0.25 & 53\% \\
RFC & Traditional & +0.49 & +0.30 & 61\% \\
\midrule
n\_prompts & NL & +0.44 & +0.23 & 52\% \\
inject\_surf & CNInterface & +0.44 & +0.27 & 61\% \\
n\_attrs & Code & +0.51 & +0.33 & 65\% \\
P\_ev & NL & +0.33 & +0.22 & 67\% \\
P\_dec\_ratio & NL & +0.36 & +0.27 & 75\% \\
n\_mem\_refs & Code & +0.50 & +0.40 & 80\% \\
n\_llm\_calls & Code & +0.38 & +0.38 & 100\% \\
\bottomrule
\end{tabular}
\end{table}

\begin{figure}[t]
\centering
\includegraphics[width=\columnwidth]{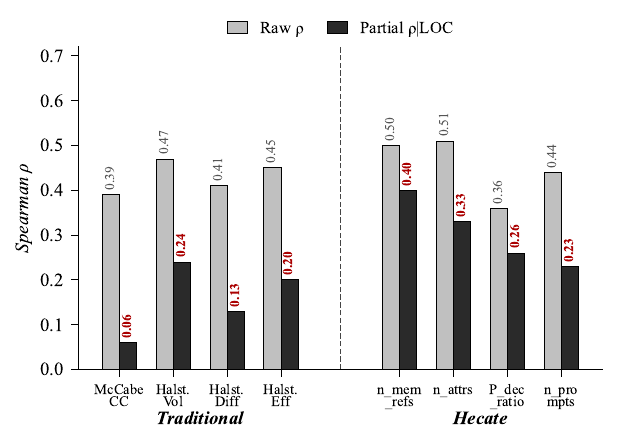}
\caption{Size-control effect on traditional (left) vs.\ \toolname{} metrics (right).}
\label{fig:rq2}
\end{figure}

Table~\ref{tab:comparison} and Figure~\ref{fig:rq2} reveal the size-confound effect.
Before size control, both traditional and proposed metrics show moderate-to-strong raw correlation with complexity.
After controlling for LOC~\cite{elemam2001confounding,zhou2018far}, the picture diverges.

\paragraph*{Why McCabe collapses}
McCabe CC drops from +0.39 to +0.06 (not significant), reproducing El Emam et al.'s~\cite{elemam2001confounding} finding that cyclomatic complexity is ``effectively a proxy for size.'' McCabe CC correlates with LOC at $r = 0.87$; branch count grows with code length, so controlling LOC removes nearly all signal.

\paragraph*{Why RFC survives}
RFC counts the set of distinct methods reachable from a class, namely own methods plus methods they call~\cite{chidamber1994metrics}. This is a coupling metric that captures how many functional units a component coordinates, which varies independently of code length. Two components of equal LOC can have RFC\,=\,5 or RFC\,=\,30 depending on architectural decomposition. RFC survives for the same reason our proposed metrics do, since it counts distinct entities rather than aggregate volume, confirming the structural-breadth principle.

\paragraph*{Why Halstead partially survives}
Halstead~N (+0.25) and Halstead~V (+0.24) are significant on all three dimensions. Although Halstead correlates with LOC at $r = 0.93$~\cite{gil2017correlations,menzies2007data}, the residual signal after size control reflects token-level information that pure line counts do not capture. The surviving Halstead signal is real but modest compared with both RFC (+0.30) and our top proposed metrics.

\paragraph*{Proposed metrics outperform}
All 7 proposed metrics of \toolname{} remain statistically significant, namely n\_mem\_refs (+0.40), n\_llm\_calls (+0.38), n\_attrs (+0.33), inject\_surf (+0.27), P\_dec\_ratio (+0.27), n\_prompts (+0.23), and P\_ev (+0.22). The two strongest (n\_mem\_refs and n\_llm\_calls) outperform every traditional survivor including RFC.

\paragraph*{Signal retention under size control}
Table~\ref{tab:comparison} reports the percentage of raw signal each metric retains after controlling for LOC. The proposed metrics retain on average 71\% of their raw correlation, compared with 43\% for the traditional baselines. n\_llm\_calls retains 98\%. Its signal is almost entirely independent of code size, because the number of LLM call sites is an architectural choice rather than a side effect of code length. By contrast, McCabe CC retains only 16\%.

\paragraph*{NL-layer independence}
To verify that NL metrics are not merely proxies for code coupling, we additionally control for n\_attrs (our strongest code metric) alongside LOC.
The NL metrics retain significant effects, namely P\_dec\_ratio (+0.27), n\_prompts (+0.29), and P\_ev (+0.21).
This confirms that the prompt layer carries maintenance-complexity information that code-only metrics cannot capture.

\answerrq{2}{After controlling for code size, traditional complexity metrics largely lose their predictive power, with RFC as the only strong survivor. All seven metrics proposed by \toolname{} remain significant, and the two strongest outperform every traditional baseline. The NL-layer metrics also remain significant after controlling for both LOC and code complexity, showing that prompt complexity contributes complementary maintenance signals.}

\subsection{Generalization to Unseen Repositories (RQ3)}\label{sec:oos}

To assess generalizability, we evaluated \toolname{} on 20 components from six repositories (crewAI, deer-flow, mem0, OpenHands, khoj, and openai-agents-python) that were excluded from the main study. The repositories were collected using the same protocol as the training set, and component extraction, quality filtering, and ground-truth mining followed the identical pipeline.

The held-out set comprises 20 components (median LOC = 939) with re-fix recurrence ranging from 0 to 73. Four proposed metrics remain positively associated with maintenance complexity after controlling for LOC: n\_attrs (+0.44), n\_mem\_refs (+0.40), P\_ev (+0.19), and P\_dec\_ratio (+0.19). The two strongest metrics from the main study, n\_attrs and n\_mem\_refs, retain nearly identical effect sizes, indicating stable generalization across repositories.

In contrast, interface-specific metrics transfer less well. Newer frameworks such as crewAI and OpenHands encapsulate prompt injection and LLM invocation behind higher-level APIs, reducing the effectiveness of API-pattern-based measurements. Traditional baselines also fail to generalize: McCabe CC and the Halstead metrics become negative after size control, while RFC retains only a weak positive effect (+0.08), consistent with the main study.

\answerrq{3}{The proposed metrics generalize to unseen repositories. The strongest metrics (\texttt{n\_attrs} and \texttt{n\_mem\_refs}) retain stable positive effects after size control, while most traditional complexity metrics again lose predictive power. Metrics tied to specific LLM APIs transfer less well because newer frameworks increasingly encapsulate prompt construction and model invocation behind higher-level abstractions.}

\section{Discussion}
\label{sec:discussion}

\subsection{Beyond Code Complexity}

Our results show that complexity in LLM-integrated applications extends beyond source code. Existing metrics measure only implementation complexity, whereas \toolname{} captures complexity in prompts, code, and their interfaces. The strongest metrics consistently measure \emph{structural breadth}, such as the number of LLM invocation sites, prompt templates, and behavioral decisions, rather than simple volume. This suggests that architectural diversity, not artifact size, is the primary driver of maintenance complexity in LLM-integrated applications.

\subsection{Implications for Engineering Practice}

Because all metrics are computed statically without executing the application or invoking an LLM, \toolname{} can be integrated into existing development workflows as a lightweight complexity gate. Project-specific thresholds can identify components whose prompt or orchestration logic is becoming increasingly difficult to maintain, complementing traditional code-quality analysis. More broadly, our findings suggest that future software metrics for AI systems should measure architectural structure rather than artifact size alone.

\subsection{Scope}

\toolname{} measures static specification complexity rather than runtime behavior. Dynamic mechanisms such as prompt construction, retrieval augmentation, or agent interactions remain outside its scope. Although our implementation targets Python, the metric definitions and validation methodology are language independent and can be extended to other LLM application frameworks with appropriate front-end support.

\section{Threats to Validity}\label{sec:threats}

\textit{Construct and internal validity.} Our three ground-truth dimensions are VCS-derived proxies rather than direct measures of complexity, although each is supported by prior metric-validation studies~\cite{basili1996validation,tornhill2015code,bird2011dont}. The component extractor achieves 87--93\% precision on a spot check, but recall is not formally evaluated, and several repositories required custom extraction rules. In addition, our NL metrics identify decision markers using keyword matching, which may miss implicit control flow.

\textit{External validity.} Our evaluation uses 118 components from 18 open-source repositories, most of which are written in Python. Although this corpus is comparable to or larger than those used in classical software-metric validation studies~\cite{chidamber1994metrics,basili1996validation,elemam2001confounding}, the findings may not fully generalize to enterprise applications or other programming languages. We therefore treat the evaluation on 20 held-out components as preliminary evidence of generalization.

\section{Related Work}\label{sec:related}

\paragraph*{Software complexity metrics}
Classical software complexity metrics, including McCabe~\cite{mccabe1976complexity}, Halstead~\cite{halstead1977elements}, and the CK suite~\cite{chidamber1994metrics}, have been widely studied as defect predictors~\cite{basili1996validation,briand2000exploring}. Subsequent work showed that many lose predictive power after controlling for size~\cite{elemam2001confounding,subramanyam2003empirical,gyimothy2005empirical,zhou2006empirical}, establishing size-controlled validation as the standard methodology. We adopt this methodology for LLM-integrated applications, where behavioral logic is distributed across both code and natural-language prompts.

\paragraph*{LLM-integrated applications}
Recent work has characterized the architecture~\cite{wang2026scaffold}, prompt engineering~\cite{promptdebt2025,parnin2025prompts}, and code quality~\cite{agentsmells2026} of LLM-integrated applications. These studies provide qualitative taxonomies but do not define or validate quantitative complexity metrics. \toolname{} fills this gap by deriving measurable metrics from these taxonomies and evaluating them using established software-metrics methodology.

\paragraph*{Prompt analysis and NL specifications}
Prompt analysis has largely focused on instruction following~\cite{complexbench2024}, prompt defects~\cite{promptdefects2025}, and tool integration~\cite{toolintegration2025}. Separately, Hoare-style reasoning has been extended to natural language for program verification~\cite{bouras2025hoareprompt}. Inspired by this line of work, our Prompt-as-Specification model treats prompts as behavioral specifications rather than verification artifacts, enabling quantitative complexity measurement across both prompts and source code. To our knowledge, \toolname{} is the first tool to measure and validate complexity jointly over the NL and code layers of LLM-integrated applications.

\section{Conclusion}\label{sec:conclusion}

In this work, we presented \toolname{}, the first static-analysis tool for measuring complexity across both prompts and source code in LLM-integrated applications. Guided by \emph{Prompt-as-Specification}, \toolname{} derives complexity metrics that capture structural properties beyond program size. Our evaluation shows that these metrics consistently outperform traditional software complexity measures and generalize to unseen repositories. Because \toolname{} is lightweight, deterministic, and requires no program execution, it can be integrated into existing CI pipelines as a complexity gate for LLM-integrated software. Future work includes extending \toolname{} to additional languages, multi-agent systems, and dynamic complexity measures.

\section*{Data Availability and Acknowledgments}

The replication package is available at \url{https://anonymous.4open.science/r/complex_artifact-868B}. 

ChatGPT and Claude have been used to assist with prose polishing during the writing of this paper.
\clearpage

\balance

\bibliographystyle{IEEEtran}
\bibliography{references}

\end{document}